\documentclass[twoside,11pt]{article}

\usepackage{lmodern}			
\usepackage{amsmath}
\usepackage{amssymb}
\usepackage{amsthm}
\usepackage[latin1]{inputenc}
\usepackage[T1]{fontenc}      
\usepackage{geometry}  
\usepackage{array}   
\usepackage[normalem]{ulem}    
\usepackage[english]{babel}		
\usepackage{hyperref}
\usepackage{color}

\newtheorem{theorem}{Theorem}

\newtheorem{definition}[theorem]{Definition}
\newtheorem{proposition}[theorem]{Proposition}
\newtheorem{corollary}[theorem]{Corollary}
\newtheorem{lemma}[theorem]{Lemma}
\newtheorem{remark}[theorem]{Remark}

\newtheorem*{theorem*}{Theorem}
\newtheorem*{hyp*}{Hypothesis}
\newtheorem*{defn*}{Definition}
\newtheorem*{prop*}{Proposition}
\newtheorem*{cor*}{Corollary}
\newtheorem*{lem*}{Lemma}
\newtheorem*{rem*}{Remark}
\newtheorem*{exe*}{Example}
\newtheorem*{todo*}{Idea(s) to check}

\DeclareMathOperator*{\argmin}{arg\,min}

\DeclareMathOperator*{\diam}{diam}
\let\refOld\ref
\renewcommand{\ref}[1]{(\refOld{#1})}

\renewcommand{\phi}{\varphi}
\newcommand{\eps}{\varepsilon}
\newcommand{\R}{\mathbb{R}}

\usepackage{amsthm}

\usepackage{hyperref}
\usepackage{amssymb}
\usepackage{amsmath}
\usepackage{natbib}

\newcommand{\comment}[1]{\color{red}\endnotetext{}\color{black}}
\newcommand{\newcomment}[1]{}
\newcommand{\verif}[1]{}

\begin{document}

\title{Recovering metric from full ordinal information}
\author{Thibaut Le Gouic}

\maketitle

\begin{abstract}

Given a geodesic space $(E,d)$, we show that full ordinal knowledge on the metric $d$ - i.e. knowledge of the function
\[
D_d:(w,x,y,z) \mapsto \mathbf{1}_{d(w,x)\leq d(y,z)},
\]
determines uniquely - up to a constant factor - the metric $d$.
For a subspace $E_n$ of $n$ points of $E$, converging in Hausdorff distance to $E$,
we construct a metric $d_n$ on $E_n$, based only on the knowledge of $D_{d}$ on $E_n$ and establish an upper bound of the Gromov-Hausdorff distance between $(E_n,d_n)$ and $(E,d)$.
\end{abstract}

\section{Introduction}

Given a set of unknown points that are known to belong to $\mathbb{R}^k$ and for which pairwise distances are known, it is useful to be able to find an embedding of these points in $\mathbb{R}^k$.
Methods to find this embedding are known as multi-dimensional scaling (MDS) methods, and are widely used as data visualization tools, in particular in social sciences.
\cite{torgerson1952multidimensional} is often considered as a pioneer paper in MDS.

This method requires the knowledge of distance between each pair of points.
We are interested in the case where only ordinal information on the distance is known.
Namely, for any four points $w,x,y,z$ in the dataset, their distance $\|w-x\|,\|y-z\|$ are not known, but they can be compared:
\[
\mathbf{1}_{\|w-x\|\leq \|y-z\|}
\]
is known.
It is a typical case in social sciences and there exist methods developed in this context.
This problem is referred as non-metric MDS or ordinal embedding.

\cite{shepard1962analysisa} and \cite{shepard1962analysisb} introduced non-metric MDS techniques, allowing to find an embedding of the data in $\mathbb{R}^{n-1}$ given a data set of $n$ points.
\cite{kruskal1964multidimensional} introduced a procedure to obtain the best possible representation in a $k$-dimensional space, for a given $k<n$.
These techniques are now widely used in practical applications, to visualize data.

The book \cite{young1987multidimensional} deals with these methods and some applications.

\subsection{Previous results in Euclidean space}

Theoretical guarantees on these methods has not been studied until recently.
Is it guaranteed that there exists a unique embedding for the data?
Given that the dataset grows up to filling a subset of the space $\mathbb{R}^k$, does the embedding of the dataset converges to the limit subset?

More formally, let $f$ be a function defined on $E\subset\mathbb{R}^k$ onto $\mathbb{R}^k$ that preserves the order of distances, i.e. such that for all $w,x,y,z\in\mathbb{R}^k$,
\begin{equation}\label{eq:isotonic}
\|w-x\|\leq \|y-z\| \text{ if and only if } \|f(w)-f(x)\|\leq \|f(y)-f(z)\|.
\end{equation}
Such functions are said to be \emph{isotonic}.
Note that there exist different variants for the definition of an isotonic function (with an "only if" instead of "if and only if", with strict inequality instead of large inequality), however, we will not discuss this into more details.
We chose what seemed the most convenient definition for this first work.

Thus, $f$ is embedding the dataset $E$ into $\mathbb{R}^k$ and preserves the ordinal information on the distances that is known.
What kind of such functions exists?
Clearly $f$ needs not to be the identity function; any similarity function (i.e. such that there exists $C>0$ such that for any $x,y\in\mathbb{R}^k$, $\|f(x)-f(y)\|=C\|x-y\|$) can fit.
A first question is then: are similarity functions the only functions that satisfy (\ref{eq:isotonic})?
We refer to this as the uniqueness question.

Given $E_n$ a set of $n$ points in $\mathbb{R}^k$ for which only $\mathbf{1}_{\|w-x\|\leq \|y-z\|}$ is known for any given points $w,x,y,z\in E_n$ - the distances between the points are unknown.
Does any function $f:E_n\rightarrow \mathbb{R}^k$ that satisfies (\ref{eq:isotonic}) satisfies that the limit of $f(E_n)$ is the limit $E$ of $E_n$ (up to a similarity)?
This refers to the consistency question.

\cite{Luxburg} provides a positive answer to both uniqueness (up to a similarity) and consistency.

The rate of convergence of the dataset embedding to its limit is tackled in \cite{Ery2015}, when the limit set $E$ is a bounded connected open set.
It essentially states that the rate of convergence of $E_n$ to $E$ is the same as $f(E_n)$ to $f(E)$ in Hausdorff metric, up to a constant factor that grows with the dimension $k$.
Methods developed in \cite{Luxburg} and \cite{Ery2015} use the vector space structure of $\mathbb{R}^k$.

The aim of this paper is to provide similar results in non Euclidean spaces.
However, since arbitrary metric spaces can not be isometrically embedded in a finite-dimensional space, we can not proceed in the exact same way to define uniqueness and consistency.

\subsection{Non Euclidean setting}

This investigation is motivated by the use of such type of information in manifold learning.
Unweighted $k$-nearest neighbor methods are widely used and fits in the framework where only a partial ordinal information on distances is known.
For instance, the ISOMAP method introduced in \cite{tenenbaum2000global} aims to learn a non linear manifold from $k$-nearest neighbor weighted graph.
Such graphs encodes partial information, much less than the full ordinal information we consider in this paper.
Little is known on what can be inferred from unweighted $k$-nearest neighbor graphs.
Our work aims to understand what can be inferred in the \emph{easier} setting of known \emph{full} ordinal information.

\medskip

In order to consider the problem for non Euclidean spaces, we make the following remark for the $\mathbb{R}^k$ case.
If $f$ is a similarity function with factor $C$ then, $E$ and $f(E)/C$ are isometric.
Stated differently, $E$ and $f(E)$ can be rescaled to have the same diameter, and then be isometric.
In particular, their Gromov-Hausdorff distance is zero.
Since Gromov-Hausdorff distance amounts to the smallest distance among all joint embedding of the two sets in a metric space (see Definition \ref{def:Haus}), this will be our way to treat the problem of ordinal embedding for non Euclidean space.

Let us formalize the problem.

\medskip

Given a metric space $(E,d)$ for which only the function
\[
D_d:(w,x,y,z) \mapsto \mathbf{1}_{d(w,x)\leq d(y,z)}
\]
is known (in other words, the metric itself is unknown but two distances can be compared), is it possible to recover the metric $d$?

The answer is clearly \textit{no} when the problem is formulated this way, because multiplying the metric by a constant does not change the known function $D_d$ (just like space can be reconstructed only up a to similarity in $\mathbb{R}^k$).
More importantly, given a sub-additive positive function $l$, such that $l(x)=0 \Leftrightarrow x=0$, then the composed function $l \circ d$ is a metric that also gives the same observed function:
\[
D_d=D_{l \circ d}.
\]

Note that this issue arises from our setting for metric space and does not happen in the setting of Euclidean space, since the embedding are always considered in the Euclidean space: a particular geodesic space (see Definition \ref{def:geod}).

Indeed, if $(E,d)$ is a geodesic space, then $(E,l\circ d)$ is geodesic only if $l$ is a \emph{positive} linear function (i.e. if $f:x\mapsto cx$ for some $c>0$).
Thus, if the space $(E,d)$ is known to be geodesic, the latter argument fails.

\subsection{Plan of the paper}

The paper falls into the following parts.

We first show that the result of uniqueness in \cite{Luxburg} holds for our setting on geodesic spaces, that is $D_d$ determines $d$ up to a constant factor.

Secondly, we present our main result which provides a construction of an estimate of a metric on a finite subspace $E_n$ of $E$ that is known to converge in Gromov-Hausdorff metric to $E$, when only $D_d$ is known on $E_n$.
Upper bounds of this convergence are proven.
The ability to build an effective estimate of the metric is a new feature on these kind of results.

Then, statistical applications are developed.

Proofs of the results follow.

\section{Uniqueness of the metric}

In order to set the problem properly, recall the definition of a geodesic space.

\begin{definition}\label{def:geod}
Let $(E,d)$ be a complete metric space.
If for any $x,y\in E$, there exists $z\in E$ such that 
\[
d(x,z)=d(y,z)=\frac{1}{2}d(x,y),
\]
then $(E,d)$ is said to be a geodesic space.
And $z$ is called a middle points of $(x,y)$ (or of the segment $[x,y]$).

A segment $[x,y]$ is a subset of $E$ such that there exists a continuous mapping $\gamma:[0,1]\rightarrow E$ such that $\gamma([0,1])=[x,y]$ and for all $t\in [0,1]$,
\[
d(x,\gamma(t)) = td(x,y) \text{ and } d(\gamma(t),y)=(1-t)d(x,y).
\]

\end{definition}

Examples of geodesic spaces that are not subset of Euclidean spaces are numerous.
Let us mention the Wasserstein space over (a subset of) $\R^k$ as an example which draws a lot of interest in statistics and machine learning (see \cite{peyre2017computational} for instance), probability and geometry (\cite{villani2008optimal} is a reference book on this subject) for example.

Our first result can then be stated as follows: the metric of a geodesic space is determined by ordinal information on the metric up to a constant factor.

\begin{theorem}\label{theorem:isometry}
Let $(E_1,d_1)$ and $(E_2,d_2)$ be two complete geodesic spaces such that there exists a surjective map $f$ such that 
\begin{equation}\label{eq:hyp}
D_{d_1}=D_{d_2\circ (f\times f)},
\end{equation}
then, there exists $c>0$ such that $f$ is an isometry between $(E_1,d_1)$ and $(E_2,cd_2)$.
\end{theorem}

Remark that as stated in the introduction, the requirement of the metric space being geodesic, can not be just dropped.
However, it is not the weakest assumption possible.
Following the steps of the proof, it is clear that it is enough to require that $(E,d)$ is a metric space such that there exist two points, joined by a shortest path, whose length is arbitrarily close to the diameter of $E$.

Note that if $(E,d)$ is a subset of $\R^k$, then it is a geodesic space if and only if this subset is convex.
The result of uniqueness of \cite{Luxburg} holds for a wider range of such subsets: uniqueness holds if $E$ is an open connected subset.
However, their results deals with embedding in $\mathbf R^k$ of a subset that inherit the distance of a geodesic space.

\begin{proof}[Proof of Theorem \ref{theorem:isometry}]
We first show that the result is true when $E_1$ is restricted to any segment $[w,x]$.

Let $w,x \in E_1$, then since $E_1$ is geodesic, there exists a middle point $m$, so that
\[
d_1(w,m)=d_1(m,x)=\frac{1}{2}d_1(w,x).
\]
Since 
\[
1=D_{d_1}(w,m,m,x)=D_{d_2\circ f\times f}(w,m,m,x)=1, \text{ and } 1=D_{d_1}(x,m,m,w)=D_{d_2\circ f\times f}(x,m,m,w)=1
\]
then,
\[
d_2(f(w),f(m))=d_2(f(m),f(x)).
\]
Thus, in order to show that $f(m)$ is a middle point of $[f(w),f(x)]$, is suffices to show that for any $m'$ such that $f(m')$ is a middle point of $[f(w),f(x)]$, 
\[
d_2(f(w),f(m))\leq d_2(f(w),f(m')).
\]
Suppose that
\[
d_2(f(w),f(m)) > d_2(f(w),f(m')),
\]
then the equality
\[
0=D_{d_1}(w,m,w,m')=D_{d_2\circ f\times f}(w,m,w,m')=0,
\]
implies
\[
d_1(w,m') < d_1(w,m).
\]
Similarly, we can show that
\[
d_1(m',x) < d_1(m,x),
\]
which contradicts that $m$ is a middle point of $[w,x]$.

We thus showed that middle points are mapped to middle points by $f$. 

Applying this recursively on a segment $[w,x]$, we show that for any $t\in [0,1]$ of the form
\[
t=\frac{k}{2^n}
\]
with $k,n\in\mathbb{N}$, and $u_t\in [w,x]$ such that $d_1(w,u_t)=td_1(w,x)$, the following holds
\begin{equation}
f(u_t)\in [f(w),f(x)] \text{ and } d_2(f(w),f(u_t))=td_2(f(w),f(x)) \label{eq:geotogeo}.
\end{equation}

Since such $t$ are dense in $[0;1]$, the result \eqref{eq:geotogeo} holds for any $t\in [0;1]$ if $f$ is continuous.
To prove that $f$ is continuous, take a sequence $w_q\rightarrow w$ as $q\rightarrow \infty$, then there exists a sequence $s_q\rightarrow 0$ such that $s_q$ is of the form $\frac{k}{2^n}$ with $k,n\in\mathbb{N}$ and there exists $u_q$ such that $d_1(w,w_q)\le d_1(w,u_q)=s_qd_1(w,x)$.
Hypothesis on $f$ then gives
\[
d_1(w,w_q)\leq d_1(w,u_q) \implies d_2(f(w),f(w_q))\leq d_2(f(w),f(u_q))=s_qd_2(f(w),f(x))\rightarrow 0.
\]

Thus, we showed that the result holds for any segment (with eventually different constants $c$).

Take now $w,x,y,z \in E_1$ with distinct $w,x$ and set 
\[
c=\frac{d_1(w,x)}{d_2(f(w),f(x))}.
\]
We want to show that constants $c$ are the same for any other segment $[y,z]$, i.e.
\[
d_1(y,z)=cd_2(f(y),f(z)).
\]
Without loss of generality, we can suppose that $d_1(y,z)\leq d_1(w,x)$. Thus, there exists $u\in [w,x]$ such that $d_1(y,z)=d_1(w,u)=td_1(w,x)$ for some $t\in [0;1]$. This equality also provides
\begin{align*}
d_1(y,z)&=td_1(w,x)&&\\
&=tcd_2(f(w),f(x))&& \text{ by definition of }c,\\
&=cd_2(f(w),f(u))&& \text{ using (\ref{eq:geotogeo}),}\\
&=cd_2(f(y),f(z))&& \text{ using (\ref{eq:hyp}) and $d_1(y,z)=d_1(w,u)$.}
\end{align*}
Thus, $f$ is an isometry between $(E_1,d_1)$ and $(E_2,cd_2)$.
\end{proof}

\section{Construction of an estimate of the metric}

Now that we know that we can construct - up to a constant factor - a geodesic metric $d$ given $D_d$, how do we build it?

To give an answer, the problem needs to be properly posed. 

Let $E_n=\{x_1,...,x_n\}$ be a subset of a geodesic \textit{compact} space $(E,d)$ of diameter $1$.
Suppose that $(E_n)_{n\geq 1}$ converges to $E$ in Hausdorff metric in $(E,d)$. 
Can we build a metric $d_n$ on $E_n$ so that $(E_n,d_n)$ converges to $(E,d)$ in Gromov-Hausdorff distance, with $d_n$ a function of ${D_d}$?

To set the notations, let us recall definitions of Hausdorff and Gromov-Hausdorff metrics.

\begin{definition}[Hausdorff and Gromov-Hausdorff metric]\label{def:Haus}
Let $A,B$ be two subset of a metric space $(E,d)$.
The Hausdorff distance between $A$ and $B$ is defined by
\[
d_H(A,B)=\inf\{\eps>0; A\subset B^\eps,B\subset A^\eps\},
\]
where $A^\eps=\{x\in E;\exists a\in A\text{ s.t. }d(a,x)<\eps\}$.

The Gromov-Hausdorff distance between two metric spaces $(E,d_E)$ and $(F,d_F)$ is defined as
\[
d_{GH}(E,F)=\inf\{d_H(g(E),h(F)); g:E\rightarrow G, h:F\rightarrow G\text{ isometric embeddings and $G$ metric space}\}.
\]
\end{definition}

More details on these metrics can be found in \cite{bbi}.

\subsection{Main result}

The idea of the proof of theorem \ref{theorem:isometry} can be used to construct a consistent pseudo-metric on $E_n$.
We define two pseudo-metrics $d^+$ and $d^-$ on $E_n$ as follows.

\begin{definition}[Pseudo metrics on $E_n$]
Let $(E,d)$ be a complete compact geodesic space, with diameter $1$.
Set $E_n=\{x_1,...,x_n\}\subset E$.
For $a,b\in E$, define - if it exists
\begin{align*}
M_{ab}&=\{z\in E; \max(d(a,z),d(b,z))\leq d(a,b)\},\\
M_{ab}^n&=M_{ab}\cap E_n\setminus \{a,b\},\\
m_{ab}&\in\argmin\{\max(d(a,z),d(b,z)); z\in M_{ab}\},\\
m^n_{ab}&\in\argmin\{\max(d(a,z),d(b,z)); z\in M_{ab}^n\},\\
\end{align*}
and set $A^n_0=(x,y)$, where $d(x,y)=\diam(E_n)$ and then for $p\geq 1$ and $A^n_p=(a_1^n,...,a_k^n)$ - if all $m^n_{a_i^na_j^n}$ exist,
\[
A^n_{p+1}=(a_1^n,m^n_{a_1^na_2^n},a_2^n,m^n_{a_2^na_3^n},a_3^n,...,m^n_{a_{k-1}^na_k^n},a_k^n).
\]
Then, for the largest $p$ such that $A^n_p$ exists, define $c_n$ on $A^n_p\times A^n_p$ by
\[
c_n(a_i^n,a_j^n)=|i-j|2^{-p}.
\]
and for any $p\geq 1$ such that $A^n_p$ exists and for any $u,v\in E$, set
\begin{align*}
d_{n,p}^+(u,v)&=\min \{c_n(a,b);d(a,b)\geq d(u,v), a,b\in A^n_p\}\\
d_{n,p}^-(u,v)&=\max \{c_n(a,b);d(a,b)\leq d(u,v), a,b\in A^n_p\}.\\
\end{align*}
Finally, set $p_n=\max\{p\in \mathbb{N}^*; A^n_p \text{ exists}, \forall a,b\in A^n_p, d^+_{n,p}(a,b) = d^-_{n,p}(a,b)\}$.
\end{definition}

\begin{remark}\label{rem:midpt}
Given $x,y$ in a geodesic space, the set of $m_{xy}$ coincides with the set of middle points of $(x,y)$.
\end{remark}

Intuitively, the largest $A_p$ is the longest geodesic path we can "make" from $E_n$, with each point being a middle point of its neighbors on $A_p$, and both $d^+_{n,p}$ and $d^-_{n,p}$ define a "metric" by comparing distances with the ones on this longest "segment" $A_p$. Then $p$ is chosen so that $d^+_{n,p}$ and $d^-_{n,p}$ are "precise" (with a high $p$) and close enough (coincide on $A_p$).

\begin{theorem}\label{theorem:main}
Let $(E,d)$ be a complete compact geodesic space, with diameter $1$.
Set $E_n=\{x_1,...,x_n\}\subset E$.

Then, for $C_0=\frac{48}{\log 2}$,
\begin{align*}
\sup_{u,v\in E_n}|d(u,v)-d^+_{n,p_n}(u,v)|&\leq C_0  d_H(E_n,E))(1-\log d_H(E_n,E))\\
\sup_{u,v\in E_n}|d(u,v)-d^-_{n,p_n}(u,v)|&\leq C_0  d_H(E_n,E))(1-\log d_H(E_n,E))
\end{align*}
\end{theorem}

Note that $d^+_{n,p_n}$ and $d^-_{n,p_n}$ are not distances, but only pseudo-metric.
Therefore, they need to be approximated by a metric in order to prove the following result.

\begin{corollary}\label{cor:gh}
Let $(E,d)$ be a complete compact geodesic space, with diameter $1$,
and $E_n$ be a finite subset of $(E,d)$.
Then, one can construct a metric $d_n$ on $E_n$, depending only on $D_{d}$ such that
\[
d_{GH}((E_n,d_n),(E,d)) \leq 2 C_0  d_{H}(E,E_n)(1-\log(d_{H}(E,E_n))),\\
\]
where $C_0=\frac{48}{\log 2}$.
\end{corollary}

\begin{remark}
This result implies that if $E_n$ converges to $E$ in Hausdorff metric, then the constructed $(E_n,d_n)$ also converge to $(E,d)$ in Gromov-Hausdorff metric.
The hypotheses $\# E_n=n$ and $E_n\rightarrow E$ in Hausdorff metric implies that $E$ is precompact.
Since it is also closed, $E$ is compact.
To relax that hypothesis, one can assume that $E_n\cap B\rightarrow E\cap B$ for any closed ball $B$.
In that case, the result states pointed Gromov-Hausdorff convergence of $(E_n,d_n)$ to $(E,d)$.
Although, since the construction of $d_n$ uses the fact that the diameter of $(E,d)$ is $1$, its construction have to be slightly adjusted.
\end{remark}

This result has an extra logaritheoremic factor compared to the one of \cite{Ery2015}, which holds in $\mathbb{R}^k$ with a constant factor growing with $k$.
It is not clear whether the logaritheoremic factor is a consequence of the method we use, or if it is needed to obtain a result independent of $k$.
Benefits of our result is that it holds in generic geodesic spaces $(E,d)$, it is not asymptotic and that a computable way to build an estimate of the metric $d_n$ is provided.

\section{Applications to statistics}

Consider now that the points $E_n=\{X_1,...,X_n\}$ of $E$ are chosen randomly, in a i.i.d. setting.
Then, if the law of $X_i$ are smooth enough, the set $E_n$ will converge to $E$ in Hausdorff metric.
The following proposition gives a more precise statement.
It is a well-known result that for which we provide a proof for completeness.

\begin{proposition}\label{proposition}
Let $(E,d)$ be a geodesic space of diameter $1$, such that 
\[
\mathcal{N}(E,t)\leq \frac{C}{t^d}
\]
for all $t>0$, where $\mathcal{N}(E,t)$ denotes the minimal number of balls of radius $t$ to cover $E$, $C$ is a positive constant, and $d$ an integer.
Set $\mu$ a Borel probability measure on $(E,d)$ such that 
\[
\mu(B_t)\geq \frac{c}{\mathcal{N}(E,t)}
\]
for some $c>0$ and any $B_t$, ball of radius $t>0$.
Set $n\in\mathbb{N}$ and let $E_n=\{X_1,...,X_n\}$ be the set of i.i.d. random variables with common law $\mu$.
Then, there exists a constant $K$ depending only on $c$ and $C$ such that,
\[
\mathbb{E}d_H(E_n,E) \leq  K \left(\frac{\log n }{n} \right)^{1/d}.
\]
\end{proposition}

Given this random set $E_n$, and metric-comparison function $D$ on this set, our theorem \ref{theorem:main} allows us to build a metric $d_n$ on $E_n$, that converges to $(E,d)$ at a speed we can control in expectation.

\begin{corollary}\label{corollary}
Under the same assumptions as in Proposition \ref{proposition}, one can construct a metric $d_n$ on $E_n$ only based on the function 
\[
D_d:(w,x,y,z)\in E_n^4 \mapsto \mathbf{1}_{d(w,x)\leq d(y,z)}
\]
such that there exists a constant $K>0$
\[
\mathbb{E}d_{GH}(E_n,E)\leq K \left(\frac{\log n}{n}\right)^{1/d}\log n.
\]
\end{corollary}

\section{Proofs}

\subsection{Main theorem \ref{theorem:main}}
The proof of theorem \ref{theorem:main} is based on the following lemmas.

\begin{lemma}\label{lem:dcn}
In the setting of theorem \ref{theorem:main}, denote $d_H$ the Hausdorff metric, then,
\[
\forall n\geq 1, \forall  p\geq 1, \forall a,b\in A_p^n, |d(a,b)-c_n(a,b)|\leq 6pd_H(E_n,E).
\]
\end{lemma}

\begin{lemma}\label{lem:pn}
In the setting of theorem \ref{theorem:main},
\begin{equation}\label{eq:pndh}
p_n\geq \left\lfloor\frac{1}{\log 2}\left( - \log (C_0 d_H(E_n,E))-\log\left( \log(e/d_H(E_n,E))\right)\right)\right\rfloor.
\end{equation}
\end{lemma}

\begin{lemma}\label{lem:duv}
In the settings of theorem \ref{theorem:main}, for $p\leq p_n$ and any $u,v\in\cup_{n\geq 1}E_n$,
\begin{enumerate}
\item $d_{n,p}^+(u,v)\leq d_{n,p}^-(u,v) + 2^{-p},$
\item $d_{n,p}^-(u,v)\leq d_{n,p}^+(u,v).$
\end{enumerate}
\end{lemma}

\begin{lemma}\label{lem:petiteineg}
Set $u\in (0,1]$, $x\in\mathbb{R}$, and $c\geq 1$, such that
\[
x\leq \log\left(\frac{1}{cu}\right)- \log\left(1-\log(u)\right),
\]
then,
\[
cxu\leq e^{-x}.
\]
\end{lemma}

\begin{proof}[Proof of lemma \ref{lem:dcn}]
Set $\eps=d_H(E_n,E)$.

{\bf First step}

Remark \ref{rem:midpt} states that since $E$ is geodesic, for all $a,b\in E$,
\[
d(a,m_{ab})\vee d(m_{ab},b) = \frac{d(a,b)}{2}.
\]
Also, by definition of the Hausdorff metric, for all $n\geq 1$, there exists $m_n\in E_n$ such that $d(m_n,m_{ab})\leq \eps$, so that
\[
d(a,m_n)\vee d(m_n,b)\leq \frac{d(a,b)}{2} + \eps.
\]
Taking $a,b \in A_p^n$, it shows that 
\[
d(a,m^n_{ab})\vee d(m^n_{ab},b) \leq \frac{d(a,b)}{2} + \eps.
\]
Using, $d(a,b)\leq d(a,m^n_{ab})\vee d(m^n_{ab},b) + d(a,m^n_{ab})\wedge d(m^n_{ab},b)$, one can show that 
\[
d(a,m^n_{ab})\wedge d(m^n_{ab},b) \geq \frac{d(a,b)}{2} - \eps.
\]
Thus, for all $a,b\in A^n_p$,
\begin{equation}\label{eq:d2}
|d(a,m^n_{ab})-\frac{d(a,b)}{2}|\leq \eps.
\end{equation}

{\bf Second step}

We want to show recursively on $p$ that for all $p\geq 0$, setting $A^n_p=(a_1,...,a_{1+2^p})$, for all $1\leq i \leq 2^p$,
\[
|d(a_i,a_{i+1})-2^{-p}|\leq (3-2^{-p})\eps.
\]
Triangular inequality and the fact that the diameter of $E$ is $1$ show that it is true for $p=0$.
Suppose it holds true for all $0\leq p \leq q$.
Then, set $A^n_{q+1}=(b_1,...,b_{1+2^{q+1}})$.
Thus, for any odd $i$ (and similarly for $i$ even), $b_{i+1}=m^n_{b_ib_{i+2}}$, so that, using (\ref{eq:d2}) and the recurrence assumption,
\begin{align*}
d(b_i,b_{i+1})&\leq \frac{d(b_i,b_{i+2})}{2} + \eps\\
& \leq 2^{-(q+1)} + (3/2-2^{-(q+1)})\eps + \eps\\
& \leq 2^{-(q+1)} + (3-2^{-(q+1)})\eps
\end{align*}

Similarly, $d(b_i,b_{i+1})\geq 2^{-(q+1)} - (3-2^{-(q+1)})\eps$.

So that, for all $1\leq i \leq 2^p$,
\begin{equation}\label{eq:lem}
|d(a_i,a_{i+1})-2^{-p}|\leq 3\eps.
\end{equation}

{\bf Third step}

Inequality (\ref{eq:lem}) proves the lemma for $p=1$.
Suppose it is true for all $1\leq p \leq k$.
Then, take $a,b\in A_{k+1}^n=(a_1,...,a_{1+2^{k+1}})$.
\begin{itemize}
\item If $a,b \in A_k^n$, then it is already supposed to be true.
\item If $a=a_i \in A_k^n$ and $b=a_j\notin A_k^n$, with $i<j$, then $a_{j-1},a_{j+1}\in A_k^n$, so that
\begin{align*}
d(a,b) - c_n(a,b) & \leq d(a_i,a_{j-1}) - c_n(a_i,a_{j-1}) + d(a_{j-1},a_j) - c_n(a_{j-1},a_j)\\
& \leq 6k\eps + 3\eps\\
c_n(a,b) - d(a,b) & \leq c_n(a_i,a_{j+1}) - d(a_i,a_{j+1}) - c_n(a_{j+1},a_j) + d(a_j,a_{j+1})\\
& \leq 6k\eps + 3\eps\\
\end{align*}
\item If $a,b \notin A_k^n$, the same ideas lead to 
\[
|d(a,b) - c_n(a,b)| \leq 6(k+1)\eps,
\]
\end{itemize}
which concludes the proof.
\end{proof}

\begin{proof}[Proof of lemma \ref{lem:pn}]
First remark that if $A_p^n$ exists and
\[
\forall a,b\in A_p^n, |d(a,b)-c_n(a,b)|< 2^{-(p+1)}
\]
then 
\[
\forall a,b\in A^n_p, d^+_{n,p}(a,b) = d^-_{n,p}(a,b).
\]
Using lemma \ref{lem:dcn} and the fact that for any $a,b\in E_n$ such that $d(a,b)\geq 2^{-p}$, the set $M_{ab}^n$ is not empty if $d_H(E_n,E)<2^{-(p+1)}$ (as it contains the closest point of $E_n$ to $m_{ab}$), one can show recursively on $p$ that $A_p^n$ exists for any $n,p$ such that $6pd_H(E_n,E)<2^{-(p+1)}$.
Thus, lemma \ref{lem:dcn} and the remark above imply that if $6pd_H(E_n,E)<2^{-(p+1)}$, then, $p_n\geq p$.
Consequently, using lemma \ref{lem:petiteineg} ( with $u=d_H(E_n,E), x=p\log 2, c=\frac{12}{\log 2}$ ), for $C_0=\frac{12}{\log 2}$,
\[
p_n\geq \left\lfloor\frac{1}{\log 2}\left( - \log (C_0 d_H(E_n,E))-\log\left( \log(e/d_H(E_n,E))\right)\right)\right\rfloor.
\]
\end{proof}

\begin{proof}[Proof of lemma \ref{lem:duv}]
Set $n\in\mathbb{N}^*$ and $p\leq p_n$ and denote $(a_1,...,a_{2^p+1})=A_p^n$.
Take $u,v\in E_n$.
\begin{enumerate}
\item Take any $a_i,a_j\in A_p^n$ such that $d^+_{n,p}(u,v)=c_n(a_i,a_j)$, $i<j$ and
\[
d(a_i,a_j)\geq d(u,v).
\]
Then, by definition of $d^+_{n,p}(u,v)$ (as a minimum),
\[
d(a_i,a_{j-1})<d(u,v)
\]
so that 
\[
d^-_{n,p}(u,v)\geq c_n(a_i,a_{j-1}) = d^+_{n,p}(u,v) - 2^{-p}.
\]
\item First, remark that since $A_p^n$ increases with $p$, $d^-_{n,p}(u,v)$ increases with $p$ and $d^+_{n,p}(u,v)$ decreases with $p$, so that is suffices to show $d^-_{n,p_n}(u,v)\leq d^+_{n,p_n}(u,v)$.
In order to show a contradiction, suppose that there exists $u,v\in \cup_{n\geq 1}E_n$ such that $d^-_{n,p_n}(u,v) > d^+_{n,p_n}(u,v)$.
Then, there exists, $a_{i_+},a_{j_+},a_{i_-},a_{j_-}\in A_{p_n}^n$ such that
\begin{align*}
c_n(a_{i_-},a_{j_-}) &= d^-_{n,p_n}(u,v),\\
d(a_{i_-},a_{j_-}) &\leq d(u,v),\\
c_n(a_{i_+},a_{j_+}) &= d^+_{n,p_n}(u,v),\\
d(a_{i_+},a_{j_+}) &\geq d(u,v),
\end{align*}
with
\begin{align}
c_n(a_{i_-},a_{j_-})&>c_n(a_{i_+},a_{j_+})\label{eq:contrad1}\\
d(a_{i_-},a_{j_-}) &\leq d(a_{i_+},a_{j_+})\label{eq:contrad2}.
\end{align}
Thus, (\ref{eq:contrad2}) gives $d^+_{n,p_n}(a_{i_-},a_{j_-})\leq d^+_{n,p_n}(a_{i_+},a_{j_+})$.

So, using definitions of $d^+_{n,p}$ and $d^-_{n,p}$ (as maximum and minimum), and definition of $p_n$,
\[
c_n(a_{i_-},a_{j_-})\leq d^-_{n,p_n}(a_{i_-},a_{j_-}) = d^+_{n,p_n}(a_{i_-},a_{j_-}) \leq  d^+_{n,p_n}(a_{i_+},a_{j_+}) \leq c_n(a_{i_+},a_{j_+}).
\]
This contradicts (\ref{eq:contrad1}), proving that hypothesis $d^-_{n,p_n}(u,v) > d^+_{n,p_n}(u,v)$ was wrong.
\end{enumerate}
\end{proof}

\begin{proof}[Proof of Lemma \ref{lem:petiteineg}]
Condition on $x$ can be stated equivalenty as
\[
e^{-x}\ge cu(1-\log(u)).
\]
Thus, the result is a consequence of applying the condition on $x$ on both sides to the inequality
\[
1-\log(u)\ge-\log(cu(1-\log(u))).
\]
To prove this last inequality, rewritten as $g(u):=1+\log(c)+\log(1-\log(u))\ge 0$, it is enough to see that $g$ is decreasing and that $g(1)=1+\log(c)\ge 0$ since $c\ge 1\ge e^{-1}$.
\end{proof}

\begin{proof}[Proof of Theorem \ref{theorem:main}]
Set $n\in\mathbb{N}^*$ and $p\leq p_n$.
Let $u,v\in E_n$.
Using lemma \ref{lem:dcn}, 
\begin{align*}
d^+_{n,p}(u,v)&=\min \{c_n(a,b);d(a,b)\geq d(u,v), a,b\in A^n_p\}\\
&\geq \min \{c_n(a,b);c_n(a,b) + 6pd_H(E_n,E)\geq d(u,v), a,b\in A^n_p\}\\
&\geq d(u,v) - 6pd_H(E_n,E).
\end{align*}
Similarly,
\[
d^-_{n,p}(u,v)\leq d(u,v) + 6pd_H(E_n,E).
\]
Thus, lemma \ref{lem:duv} implies
\begin{align*}
d(u,v) - 6pd_H(E_n,E) - 2^{-p} &\leq d^+_{n,p}(u,v)-2^{-p} \\
&\leq d^-_{n,p}(u,v)\\
&\leq d^-_{n,p_n}(u,v)\\
&\leq d^+_{n,p_n}(u,v)\\
&\leq d^+_{n,p}(u,v)\\
&\leq d^-_{n,p}(u,v) + 2^{-p} \leq d(u,v) + 6pd_H(E_n,E)+ 2^{-p}.
\end{align*}

Taking
$p= \left\lfloor\frac{1}{\log 2}\left( - \log (C_0 d_H(E_n,E))-\log\left( \log(e/d_H(E_n,E))\right)\right)\right\rfloor\leq p_n$ as in (\ref{eq:pndh}), lemma \ref{lem:petiteineg} implies that $6pd_H(E_n,E)\leq 2^{-p-1}$, so that

\begin{align*}
\sup_{u,v\in E_n}|d(u,v)-d^+_{n,p_n}(u,v)|&\leq 2^{-p+1} \leq 4 C_0  d_H(E_n,E))(1-\log d_H(E_n,E))\\
\sup_{u,v\in E_n}|d(u,v)-d^-_{n,p_n}(u,v)|&\leq 2^{-p+1} \leq 4 C_0  d_H(E_n,E))(1-\log d_H(E_n,E))
\end{align*}
\end{proof}

\subsection{Corollary \ref{cor:gh}}

\begin{proof}[Proof of corollary \ref{cor:gh}]
It suffices to choose the closest metric $d_n$ to $d^+_{n,p_n}$ in the sup sense:
\[
d_n \in \argmin \left\{ \sup_{u,v\in E_n}|d(u,v)-d^+_{n,p_n}(u,v)|; d \text{ is a metric on } E_n\right\}.
\]
Then, since $E_n\subset E$, there exists a surjective map $f:E\mapsto E_n$ such that
\[
d_{GH}(E_n,E)\le d_H(E_n,E)=\sup_{u,v\in E}|d(u,v)-d(f(u),f(v))|
\]
so that
\begin{align*}
d_{GH}((E_n,d_n),(E,d))&\leq d_{GH}((E_n,d_n),(E_n,d)) )\\
&\leq \sup_{u,v\in E_n}|d(u,v)-d_n(u,v)| \\
&\leq 2 \sup_{u,v\in E_n}|d(u,v)-d^+_{n,p_n}(u,v)| \\
&\leq C d_H(E_n,E))(1-\log d_H(E_n,E))
\end{align*}
\textit{The argmin that define $d_n$ does not actually necessarily exists, but any metric close enough satisfies it too.}
\end{proof}

\subsection{Proposition \ref{proposition}}

\begin{proof}
For $t>0$, denotes $\mathcal{N}(E,t)$ by $m_t$. Given balls $(B_i)_{1\leq i \leq m_{t/2}}$ that cover $E$, 
\begin{align*}
\mathbb{E}d_H(E_n,E) & \leq \mathbb{E}d_H(E_n,E)\mathbf{1}_{\{d_H(E_n,E)>t\}} + \mathbb{E}d_H(E_n,E)\mathbf{1}_{\{d_H(E_n,E)\leq t\}}\\
& \leq \mathbb{P}(d_H(E_n,E)>t) + t\\
& \leq \mathbb{P}\left( \bigcup_{1\leq i \leq m_{t/2}} \bigcap_{1\leq k \leq n} \{X_k \notin B_i\}\right) +t\\
& \leq \sum_{1\leq i \leq m_{t/2}}\prod_{1\leq k \leq n} e^{\log(1-\mu(B_k))} +t\\
& \leq m_{t/2} e^{-\frac{cn}{m_t}} + t \\
& \leq \frac{2^dC}{t^d} e^{-cnt^{d}/C} + t.
\end{align*}
Choosing $t=\left(\frac{C(1+1/d)}{c}\frac{\log(n)}{n}\right)^{1/d}$ leads to
\begin{align*}
\mathbb{E}d_H(E_n,E) & \leq \frac{2^dcn}{(1+1/d)\log n} e^{-(1+1/d)\log n} + \left(\frac{C(1+1/d)}{c}\frac{\log(n)}{n}\right)^{1/d}\\
& \leq K \left(\frac{\log n }{n} \right)^{1/d}
\end{align*}

\end{proof}

\bibliography{bibli}
\end{document}